\documentclass{article}

\usepackage[preprint]{neurips_2023}
\usepackage{listings}
\usepackage[utf8]{inputenc} 
\usepackage[T1]{fontenc}    

\usepackage{url}            
\usepackage{booktabs}       
\usepackage{amsfonts}       
\usepackage{nicefrac}       
\usepackage{soul}
\usepackage{microtype}      

\usepackage{amsmath}
\usepackage{subfig}
\usepackage{CJKutf8}

\usepackage[table,dvipsnames]{xcolor}

\usepackage{tabstackengine}
\usepackage{nicematrix}
\usepackage{tabularx, booktabs}
\usepackage{tablefootnote}
\usepackage{amssymb}
\usepackage{minitoc}
\usepackage{tikz,graphics,color,fullpage,float,epsf,caption,subcaption}
\usepackage{graphicx}
\usepackage{subcaption}

\usepackage{makecell}

\usepackage{tabulary,multirow,overpic}

\definecolor{citecolor}{RGB}{34,139,34}

\usepackage[breaklinks=true,letterpaper=true,colorlinks, citecolor=citecolor,bookmarks=false]{hyperref}

\newcolumntype{x}[1]{>{\centering\arraybackslash}p{#1pt}}

\newcommand{\app}{\raise.17ex\hbox{$\scriptstyle\sim$}}

\newlength\savewidth\newcommand\shline{\noalign{\global\savewidth\arrayrulewidth
  \global\arrayrulewidth 1pt}\hline\noalign{\global\arrayrulewidth\savewidth}}

\title{Intent-based Prompt Calibration: Enhancing prompt optimization with synthetic boundary cases}
\author{%
  Elad Levi\\
  \And
  Eli Brosh\\
  \And
  Matan Friedmann
}

\begin{document}
\doparttoc 
\faketableofcontents 
\part{} 

\maketitle

\begin{abstract}
Prompt engineering is a challenging and important task due to the high sensitivity of Large Language Models (LLMs) to the given prompt and the inherent ambiguity of a textual task instruction. Automatic prompt engineering is essential to achieve optimized performance from LLMs. Recent studies have demonstrated the capabilities of LLMs to automatically conduct prompt engineering by employing a meta-prompt that incorporates the outcomes of the last trials and proposes an improved prompt. However, this requires a high-quality benchmark to compare different prompts, which is difficult and expensive to acquire in many real-world use cases. In this work, we introduce a new method for automatic prompt engineering, using a calibration process that iteratively refines the prompt to the user intent. During the optimization process, the system jointly generates synthetic data of boundary use cases and optimizes the prompt according to the generated dataset. We demonstrate the effectiveness of our method with respect to strong proprietary models on real-world tasks such as moderation and generation. Our method outperforms state-of-the-art methods with a limited number of annotated samples. Furthermore, we validate the advantages of each one of the system's key components. Our system is built in a modular way, facilitating easy adaptation to other tasks. The code is available at \href{https://github.com/Eladlev/AutoPrompt}{https://github.com/Eladlev/AutoPrompt}.

\end{abstract}

\section{Introduction}
\label{sct:introduction}
In recent years, there has been significant enhancements in the capabilities of Large Language Models (LLMs), demonstrating impressive generative performance across a variety of tasks~\cite{bert,gpt3}. Nevertheless, despite these advancements, the quality of the models' outputs is highly sensitive to the conditioned prompt~\cite{Lu2021FantasticallyOP, DBLP:conf/icml/ZhaoWFK021}. Even a slight modification in the prompt format can significantly impact the model's performance~\cite{DBLP:journals/corr/abs-2310-11324}. This issue is even more evident in popular proprietary models, where a change in model version results in drastic changes in model behaviour on a wide range of tasks~\cite{DBLP:journals/corr/abs-2307-09009}.

In order to tackle the prompt sensitivity issue, several methods~\cite{lester-etal-2021-power,li-liang-2021-prefix} proposed to use soft prompts which require access to the LLM itself in order to perform the optimization. Recently, ~\cite{OPRO,PE,pryzant-etal-2023-automatic} demonstrated the effectiveness of using LLMs themselves to optimize the prompt. To this end, each prompt is assigned a score based on a given benchmark and an appropriate metric. The optimization process is performed  iteratively by providing a meta-prompt that incorporates the history of the last few prompt scores and guiding the model to suggest a better prompt with a higher score. 
However, the high-quality, large benchmarks required by this approach to evaluate the performance of the different prompts often do not exist in  many real-world use cases. Moreover, iterating on such large datasets can be costly.

LLMs have proven to be highly effective in generating high-quality and rich datasets that boost model performance on a diverse set of tasks~\cite{DBLP:journals/corr/abs-2308-12950, DBLP:journals/corr/abs-2312-02418, DBLP:journals/corr/abs-2308-09583, DBLP:journals/corr/abs-2401-00368}. Recent works demonstrate the capabilities of LLMs to refine the prompt provided by the user, resolving the initial prompt ambiguity~\cite{DBLP:journals/corr/abs-2311-04205}. However, without additional information, the model has to guess the true intention of the user, which in many cases can lead to inaccurate results.

In this work, we introduce Intent-based Prompt Calibration (IPC), a system which aims to calibrate the prompt according to the intention of the user, by using synthetic examples. The calibration process is performed by iteratively building a dataset of challenging boundary cases and optimising the prompt according to the generated benchmark. This novel aspect of our method, producing a small benchmark tailored to the boundary cases of the user's task as part of the optimization process, is highly valuable for explainability, LLM distillation, and other use cases. In contrast to previous works, the system is optimized for real-world use cases such as moderation which usually suffers from imbalanced data distribution. We also extend the prompt optimization to a new family of generative tasks, by first fitting a ranking prompt and then performing the prompt optimization with the learned ranker. Learning a prompt ranker allows us to optimize generative tasks with minimal annotation effort. As demonstrated in our experimentation section,  using such an approach without synthetic data of boundary cases, e.g., as done in previous methods, would not be efficient due to the natural imbalance of the ranking distribution. 

Lastly, our system is built in a modular way, such that each part of the components can be used on its own in other tasks like synthetic data generation or prompt distillation between two LLMs. We describe the system components in detail and demonstrate the effectiveness of our proposed method with respect to strong proprietary models like GPT-3.5/4-Turbo. We show that our method outperforms previous methods using a very small amount of data and iteration steps . This significantly reduces the total optimization efforts and costs, and makes the system applicable to various production use cases.

\begin{figure}
  \centering
  \includegraphics[width=0.7\linewidth]{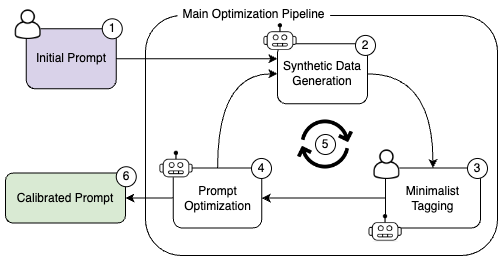}
  \caption{System diagram. (1) An initial prompt is provided by the user (2) Synthetic challenging cases are generated (3) A user or an LLM annotates the examples (4) After evaluating the prompt performances, an LLM suggest a new prompt given the last prompt's results. (5) This process is repeated iteratively until a certain stop criterion (6) The system outputs a calibrated prompt.}
  \label{fig:system_diagram}
\end{figure}

\section{Method}
\label{sct:method}
Our system is illustrated in Figure \ref{fig:system_diagram}. We start with the initial prompt suggestion and a task description. The user can also provide a few examples in a few-shot setting. Then, during the calibration optimization process, the system iteratively: 1. Suggests a few samples of challenging and diverse boundary cases for the task and the current prompt. 2. Evaluates the current prompt on the generated dataset, and provides an analysis. 3. Given the history of the last few prompts, suggests a new prompt with a higher score. The optimization process is terminated when either there is no improvement in the last few steps, or when the maximum number of iterations has been reached.

\begin{figure}
  \centering
  \includegraphics[width=0.9\linewidth]{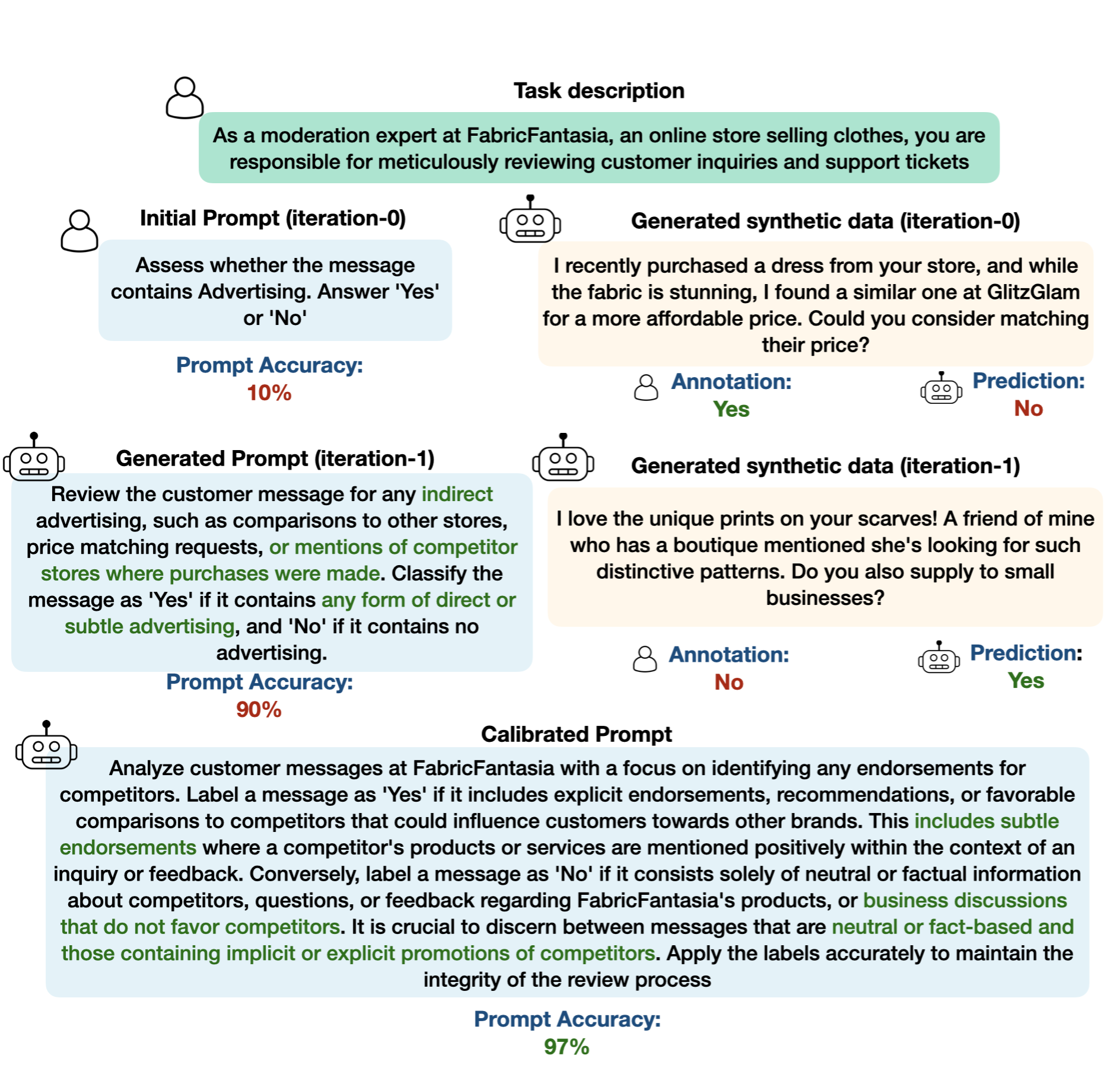}
  \caption{Example of a real system flow. The user provides only the task description and initial prompt. The model iteratively generates challenging samples and refines the prompt according to the generated benchmark.}
  \label{fig:ex_ad}
\end{figure}

The base configuration of our system is optimized for classification tasks, with accuracy set as the score function, and the error analysis determined by a confusion matrix and the prompt misclassifications. An example of the system flow can be seen in Figure \ref{fig:ex_ad}. In each iteration, new challenging samples are generated (according to the current prompt), and the misclassifications are used to refine the prompt until it is calibrated to the user intent.

\subsection{Generative tasks}
\label{sct:gen-tasks}
To extend the prompt calibration process from classification tasks to generative tasks, we split the optimization process into two parts. In the first part, we use an LLM to rephrase the initial prompt and task description in order to define a ranking task based on the modified initial prompt and task description. We then calibrate a prompt for the ranking task, treating it as a classification task using the classification pipeline. Naturally, the ranker distribution tends to be a normal distribution with its mean at the mean score. This distribution is imbalanced, especially in the interesting range of the top scores. Therefore, in the ranking case, the sample generator meta-prompt is instructed to generate challenging boundary samples from the top two scores.
In the second part, we leverage the same underlying process to optimize the original generative prompt. This step is done by iteratively applying steps 2 and 3, described in the system overview, using the calibrated ranking prompt as the score function.
It's important to note that human annotations are required only in the ranking calibration process. Furthermore, by treating the intent as a classification task, the prompt can be calibrated using a small amount of annotation effort.

\subsection{Meta-Prompts}
The meta-prompts consist of three separate prompts, as can be seen in Appendix A.

\textbf{Sample generator.} The sample generation meta-prompt is determined according to the system state: In the first iteration, if the user doesn't provide any samples (zero-shot setting), the meta-prompt instructs the model to generate diverse adversarial samples with even class distribution. In the next iterations, the prompt is extended with the following additional context: (1) A history with prompts and good adversarial samples that confused the prompts; and (2) A set of realistic samples from the dataset, where the model is instructed to preserve the dataset style.
The context-realistic samples are chosen to be semantically close according to a given sentence embedding. 

\textbf{Analyzer.} The analyzer meta-prompt receives the prompt score, a confusion matrix in the classification case, and a set of errors in all the classes. It is then instructed to produce an analysis summary of the prompt performances and the major failure cases.

\textbf{Prompt generator.} The input for the prompt generator meta-prompt is (1) A list of the last suggested prompts and their scores (2) The performance analysis of the last prompt that is produced by the Analyzer prompt. The model is instructed to produce a prompt with a higher score according to the history and the analysis.

\subsection{System pipeline}
\label{sct:sys-pipeline}
An overview of the system architecture can be seen in Figure~\ref{fig:arch}. The system consists of four primary components.

\textbf{Dataset.} This component manages the dataset and performs operations such as insertion, modification, deletion, and applying functions,  on the dataset rows. The component also handles data cleaning by removing semantic duplications and performing semantic sampling. Since the system is optimized for small datasets, the current implementation is based on a local database using pandas.

\textbf{Estimator.} The estimator is responsible for estimating a batch of samples. We implement this component twice, once for the predictions and once for the annotations. This generic implementation for both types of use cases, allows us to modify the system simply for diverse use cases such as prompt calibration, prompt distillation and prompt squashing. The currently supported types of estimators are: (1) Human annotation, using Argilla UI~\cite{argilla}. The system is connected to the Argilla server and is waiting until the annotation task is completed; (2) LLM estimator, which uses an LLM to estimate the sample given a prompt. We support various types of LLMs, using Langchain integration~\cite{langchain}. For efficiency, the system supports parallelism using both workers and async calls. The system also supports sending a few samples in one prompt (prompt batching), which can reduce the cost significantly; and (3) Batch estimator, the batch estimator runs multiple LLM estimators and integrates their outputs through an aggregation layer. It is mainly used for prompt-squashing, enabling users to optimize a single prompt that will perform as well as running few prompts multiple times. For example,  when a user wants to apply several moderation rules simultaneously.

\textbf{Evaluator.} The evaluator is responsible for evaluating the records after the prediction and annotation stage. The evaluator accepts a function and applies it to each row. It's important to note that the function is generic. For example, in the generation pipeline, the function is performed by invoking an LLM. The evaluator is also responsible for defining the errors and handling the error analysis using the Analyzer described in the meta-prompts section.

\textbf{Optimizer (Optimization Pipeline).} The optimizer manager handles the whole optimization process flow, it performs the iteration steps described in the previous section and is responsible for stopping and returning the final calibrated prompt. The currently supported criteria are either convergence (determined by a patience hyper-parameter), or usage limit (determined by maximal cost if relevant, or by the number of generated tokens).


\begin{figure}
    \centering
    \begin{minipage}[b]{0.49\linewidth}
        \centering
        \includegraphics[width=\linewidth]{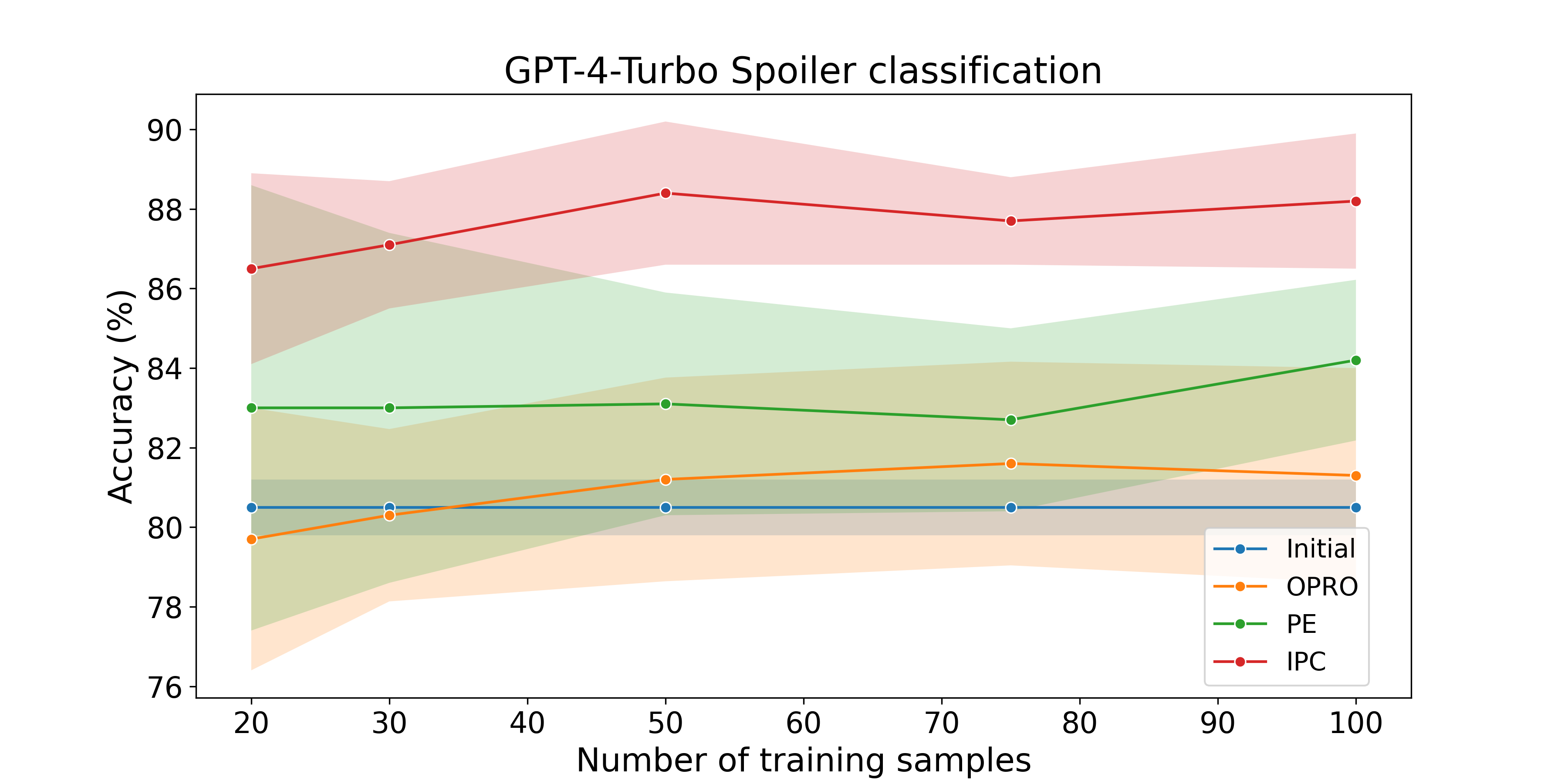}
        \label{fig:image1}
    \end{minipage}
    \hfill
    \begin{minipage}[b]{0.49\linewidth}
        \centering
        \includegraphics[width=\linewidth]{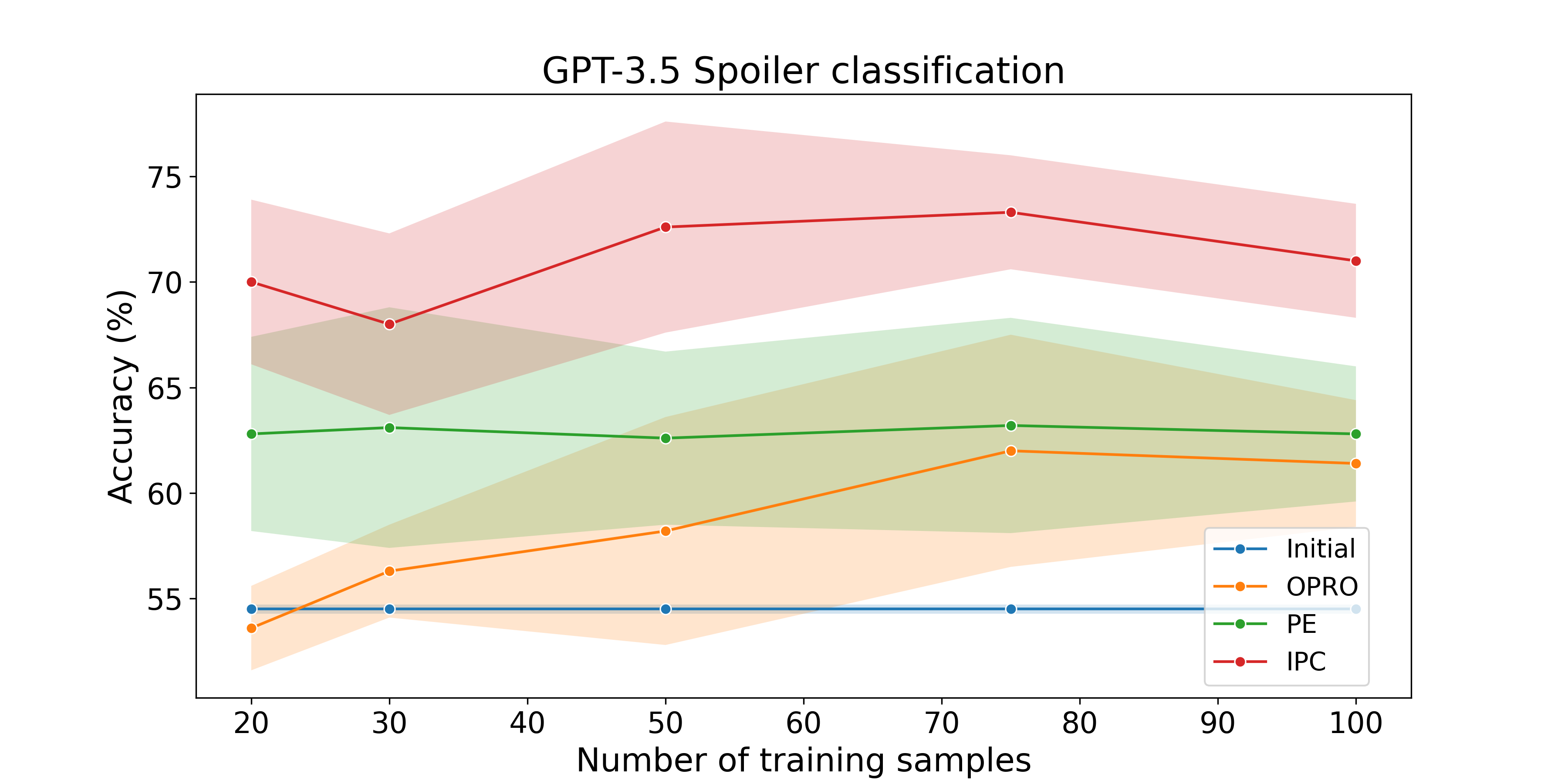}
        \label{fig:image2}
    \end{minipage}
    
    \begin{minipage}[b]{0.49\linewidth}
        \centering
        \includegraphics[width=\linewidth]{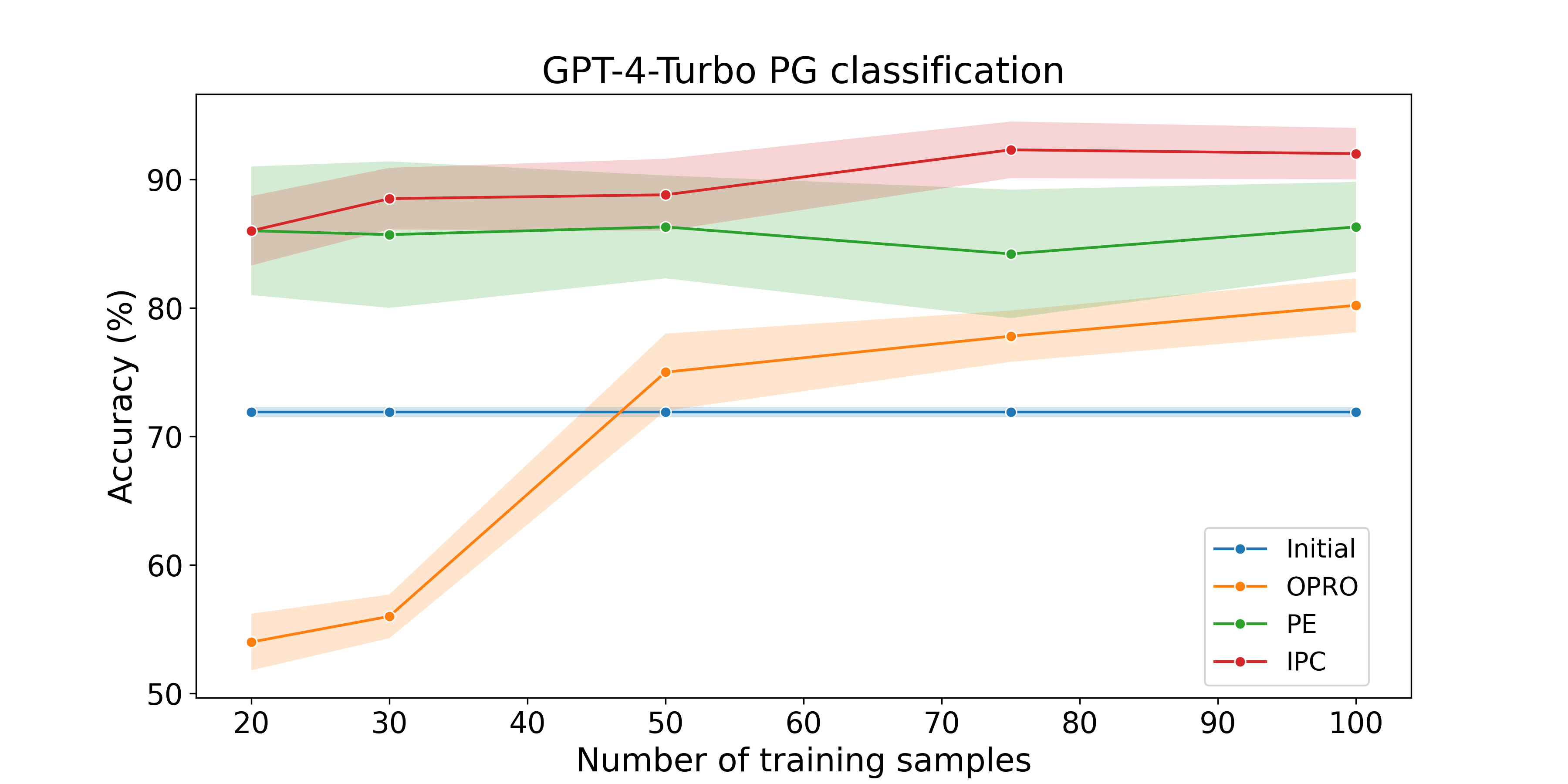}
        \label{fig:image3}
    \end{minipage}
    \hfill
    \begin{minipage}[b]{0.49\linewidth}
        \centering
        \includegraphics[width=\linewidth]{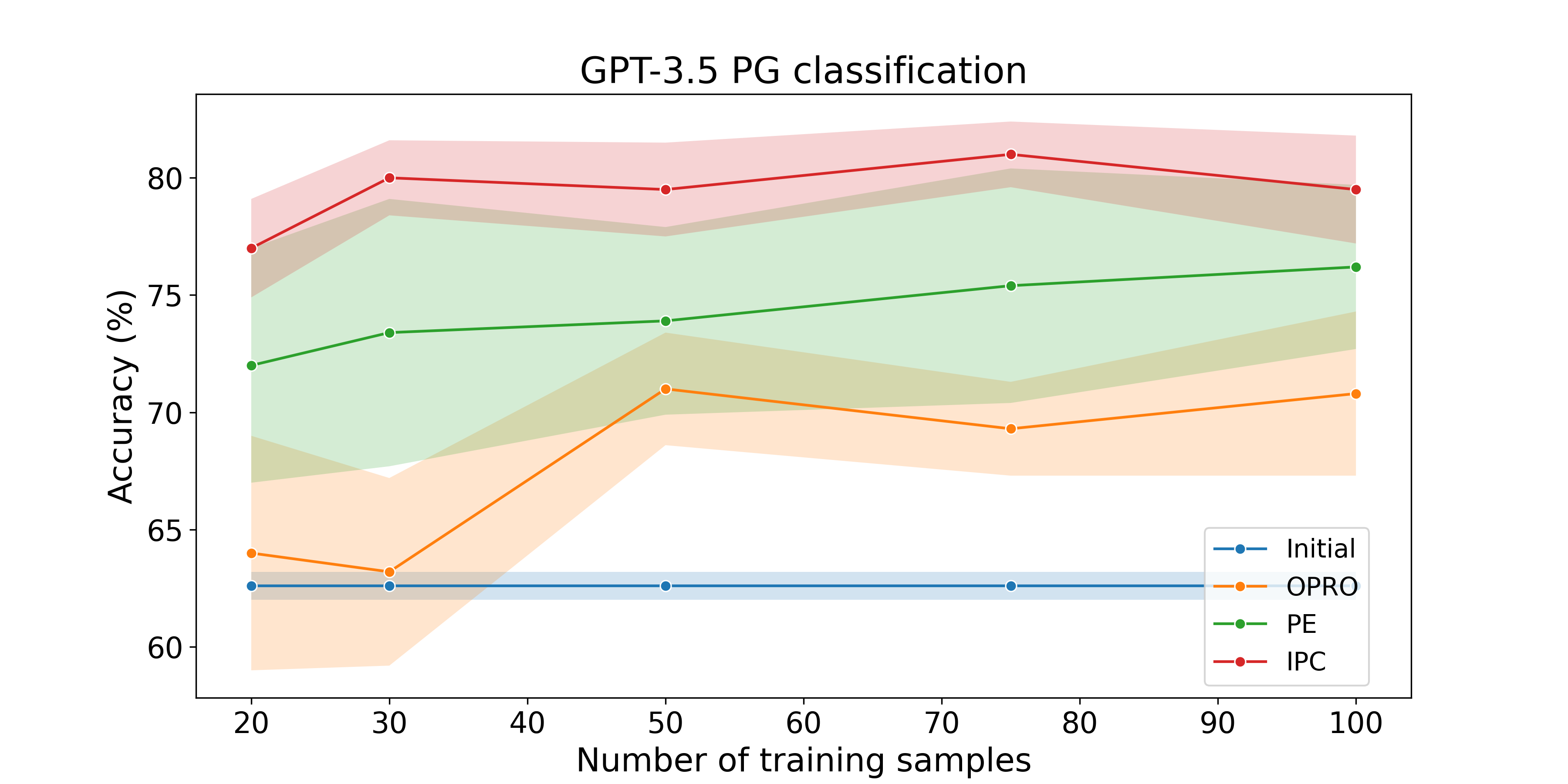}
        \label{fig:image4}
    \end{minipage}
    \label{fig:all_images}

    \caption{Accuracy on the spoiler and the PG classification tasks, with respect to the number of training steps. As shown, IPC outperforms other tested methods and results in lower variance.}
    \label{fig:spoiler}
\end{figure}

\section{Experiments}
\label{sct:exp}

We test our system on scenarios that reflect real-world moderation and generation use cases on strong proprietary models (GPT-3.5/4-Turbo). We used the IMDB review dataset~\cite{IMDB} as the base data for all our experiments. We compare our proposed IPC method to two SOTA prompt optimization methods that are based on meta-prompts: 1. OPRO \cite{OPRO} and 2. The meta prompt provided by PE \cite{PE}.
\subsection{Classification}
We evaluate the prompt calibration process on three binary classification tasks: (1) Spoiler detection, (2) Sentiment analysis, and (3) Parental Guidance (PG) detection. In each experiment, we start with some initial samples and a prompt, in addition to a short task description. To generate the ground truth (GT) we composed a highly detailed prompt with specific preferences. For the GT generation, we used a strong model (GPT-4 Turbo), as this process of generating the GT simulates a user's particular preference for the given task. The baseline methods were trained on samples from the IMDB dataset~\cite{IMDB}, whereas our proposed method was trained on the adversarial synthetic data which was provided with 10 initial samples from the original IMDB training data. All methods were trained for 50 iterations. The test dataset was taken from the IMDB reviews test split, with the generated annotations provided by the GT prompt. We then collected 250 samples for each class in each one of the tested scenarios, such that the final dataset has equal class distribution. It's important to note that the IMDB dataset includes only highly polarizing reviews (no reviews with ratings in the range of 4-7). To evaluate the method's performance on more challenging cases, we also generate a synthetic test dataset with 300 samples (using the initial prompt) for the sentiment classification task.

\begin{figure}[tp]
  \centering
    \begin{minipage}{.49\linewidth}
        \centering
        \includegraphics[width=\linewidth]{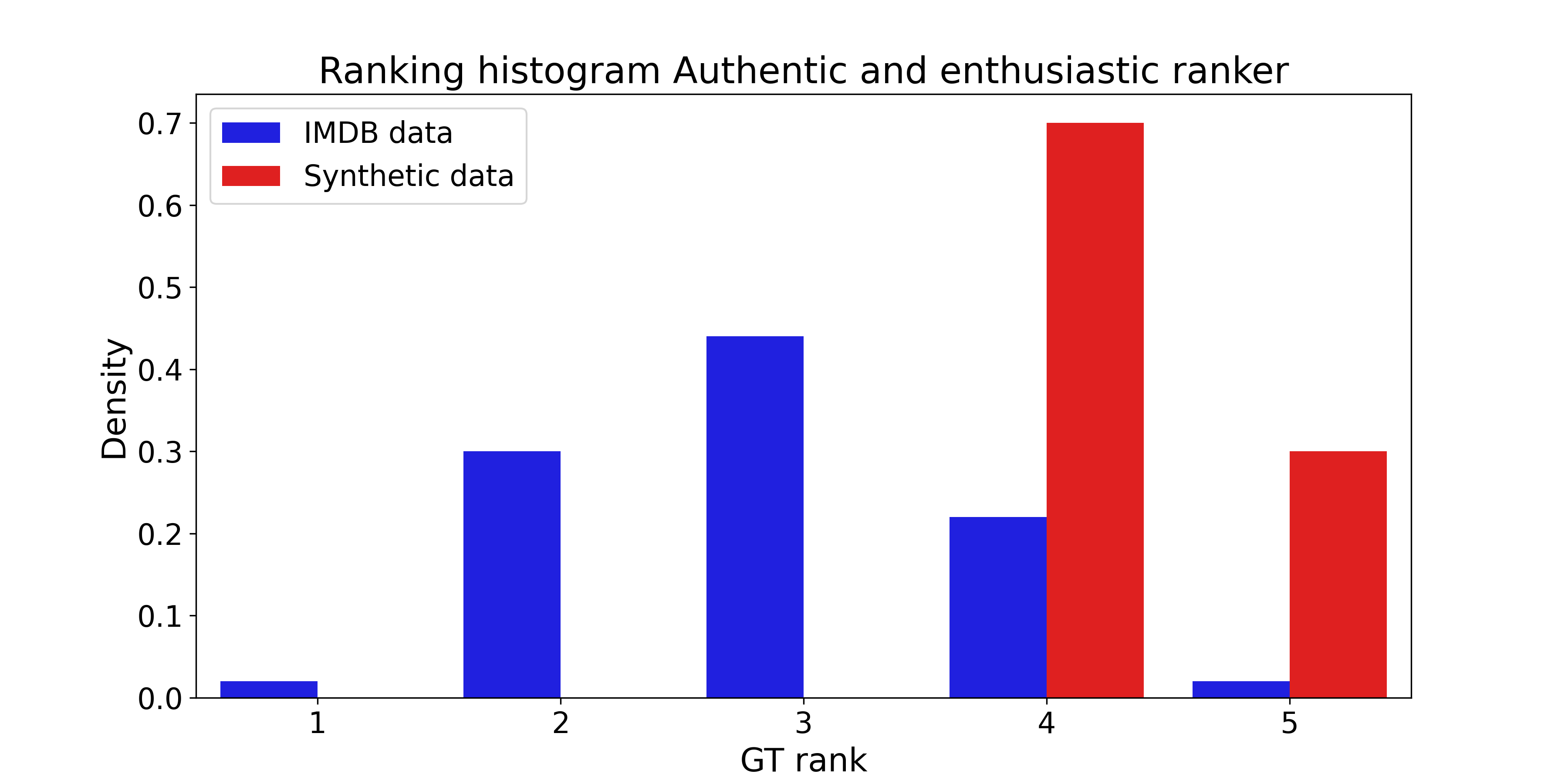}
        \caption{Histogram of the ranking scores of the IMDB review dataset vs the generated synthetic dataset on the Authentic and enthusiastic generation task. The real data distribution contains very few ranked 5 reviews, whereas the synthetic data contains a more balanced dataset with respect to the top scores.}
       \label{fig:hist_rank}
    \end{minipage} \hfill
    \begin{minipage}{.49\linewidth}
        \centering
        \includegraphics[width=\linewidth]{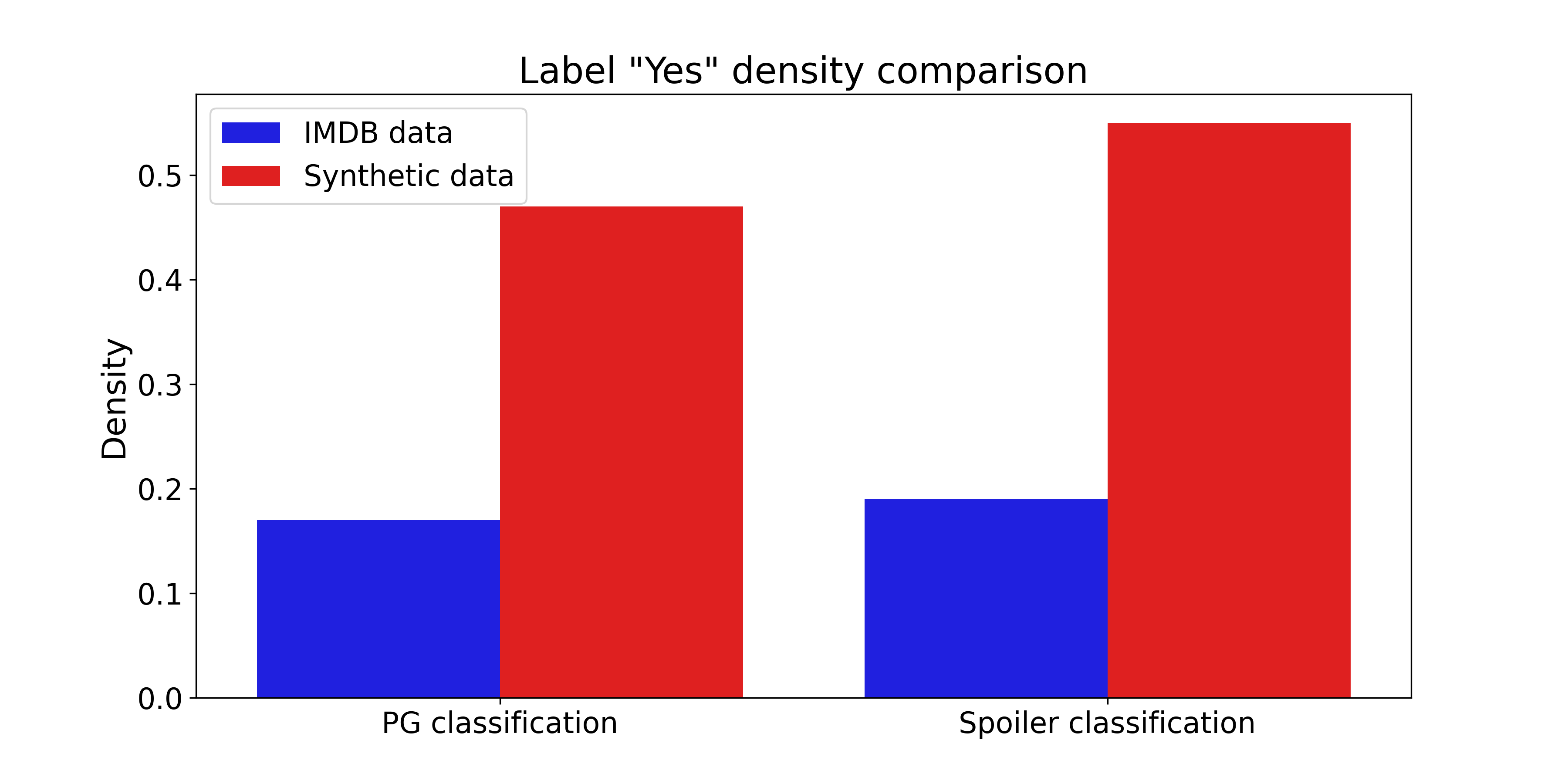}
        \caption{Histogram of the 'Yes' labels density on the parental guidance (PG) and spoiler classification tasks, with respect to the IMDB review dataset and the generated synthetic dataset. The real data exhibits a heavy imbalance in favor of the 'No' label, while the synthetic data approaches an even distribution. }
        \label{fig:moderation_hist}
    \end{minipage}
\end{figure}

We present our results in Figures \ref{fig:spoiler},\ref{fig:all_images}. As seen in the figures, IPC outperforms all other tested methods. In particular, it's important to note the high variance of the other methods, especially in the case of a small number of training samples. The gap in performance between the methods becomes even more evident in the synthetic data case, where there are more boundary cases, as can be seen in Figure \ref{fig:synt}.
A qualitative comparison between the methods can be seen in Table \ref{tab:spoiler}. While OPRO~\cite{OPRO} results in mainly rephrasing the initial prompt, and the PE~\cite{PE} prompt only partly fits the GT prompt, the IPC prompt successfully captures the subtle details and nuances of the GT prompt. The significant differences in data distributions between the original data and the synthetic generated data can be seen in Figure \ref{fig:moderation_hist}. Both the spoiler and the PG classification tasks exhibit significant bias towards the 'No' labels, where the Synthetic data is almost balanced.


\begin{figure}[tp]
  \centering
{\includegraphics[width=0.49\textwidth]{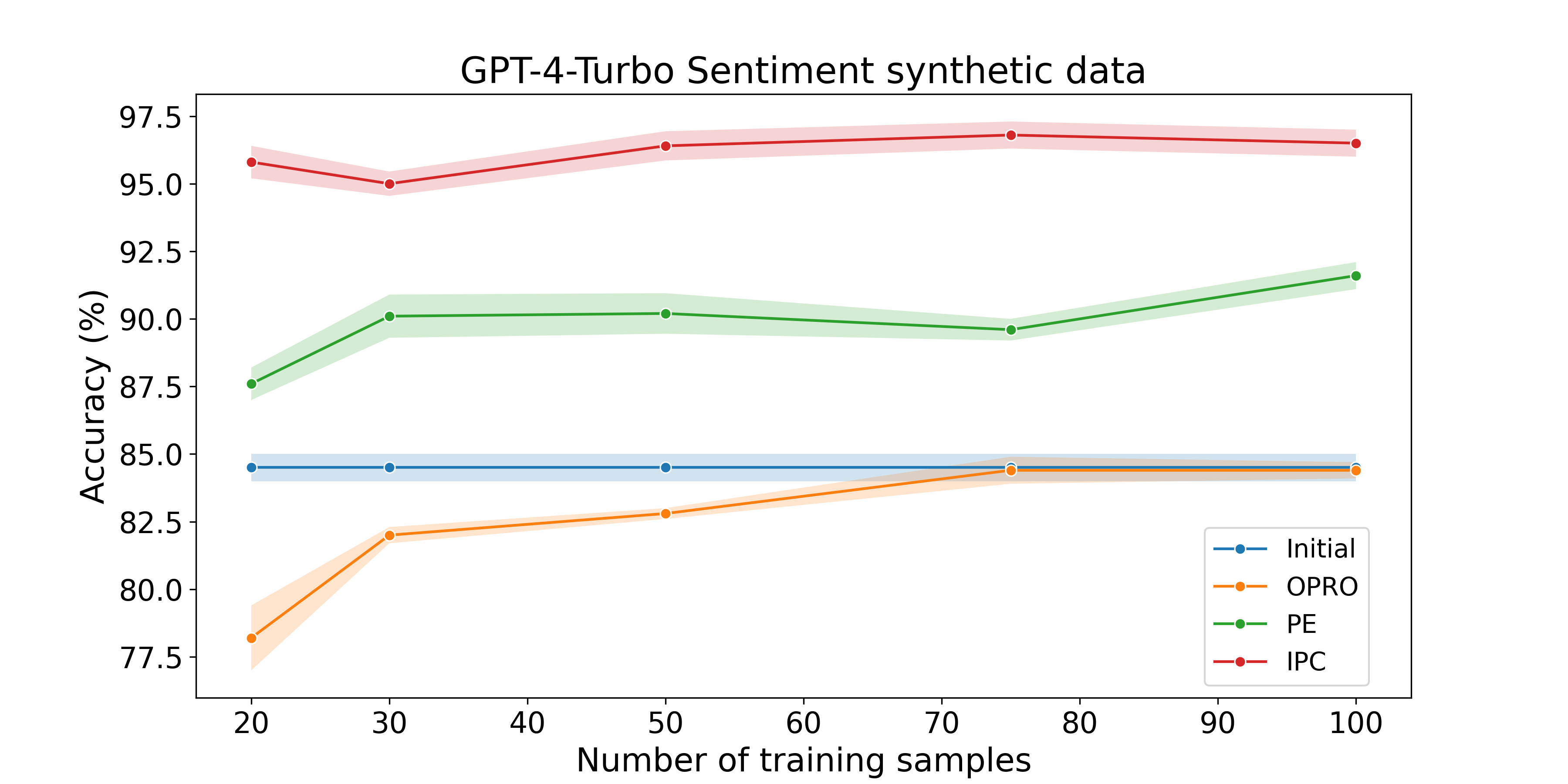}}
  \hfill
{\includegraphics[width=0.49\textwidth]{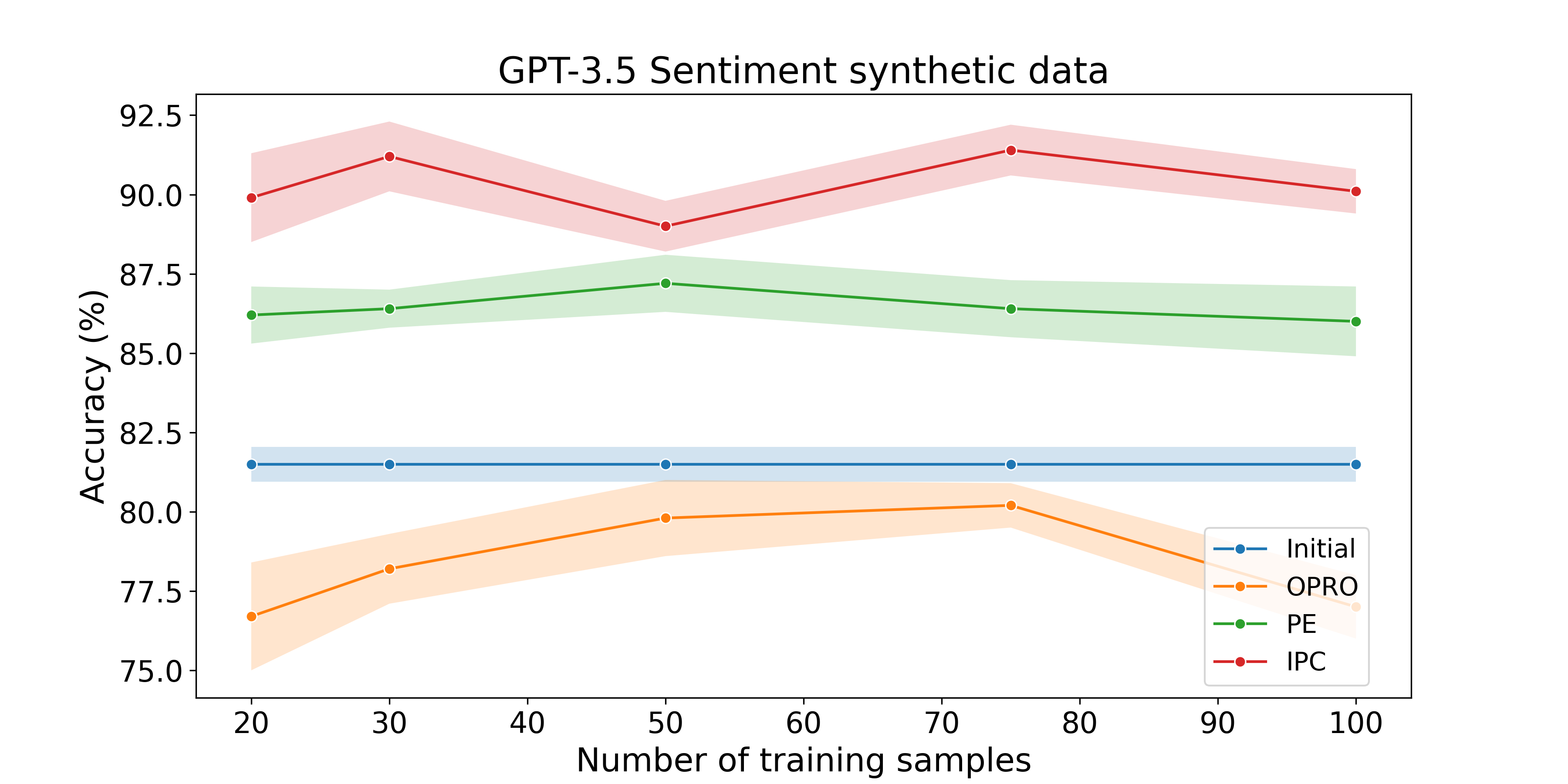}\label{fig:synt}}
  \caption{Accuracy of of sentiment classification task with respect to the synthetic dataset for different number of training steps. IPC outperforms other tested methods and results in lower variance.}
  \label{fig:synt}
\end{figure}

\begin{table*}[!h]
\begin{center}
\begin{tabular}{l| p{12cm}l} 
\shline
\\ 
 GT prompt & To improve the IMDB review spoiler classifier, label a review as 'Yes' for containing a spoiler if it: 1. Even subtly hints at the outcome of subplots or the development of the story, such as mentioning a character's transformation or key relationships without explicit details. 2. Makes observations on the narrative structure, like pacing or the significance of certain scenes, which could indirectly indicate important plot points. 3. Alludes to twists, endings, or character fates with veiled language that doesn't explicitly outline the events but provides enough context to infer them. Conversely, label a review as 'No' for no spoilers if it: 1. Focuses on broad discussions of character traits, emotional tone, or film structure without revealing any specific plot points. 2. Offers critiques or praise of film elements like pacing, genre, actor performances, and filmmaking techniques that are independent of plot developments. 3. Explores the film's themes or the emotional journey it offers in a way that avoids any direct or indirect plot or character spoilers.\\ [0.5ex] 
 \hline
 \\ 
 Initial prompt & Does this movie review contain a spoiler? answer Yes or No \\ 
 \hline
\\ 
OPRO~\cite{OPRO} & Examine the movie synopsis with vigilance for spoiler content and succinctly affirm their presence with "Yes" or negate with "No". \\
 \hline
 \\ 
 PE~\cite{PE} & Analyze the IMDB review for elements that provide significant insights into the plot or reveal crucial turning points, even if mentioned subtly or indirectly. Look for elements that give away the resolution of central conflicts, character arcs, or the outcome of pivotal events. Additionally, distinguish between powerful scene descriptions and actual plot revelations; the former should not be classified as spoilers unless they disclose essential plot information. Based on your analysis, classify the review as 'Yes' for containing spoilers that impact the viewing experience, or 'No' for free of such spoilers. \\
 \hline
 \\
 IPC & IMDB Review Spoiler Identification Protocol: For the task of classifying IMDB reviews for the presence of spoilers, the classifier must label reviews with a heightened sensitivity to nuanced language and indirect spoiler cues. The classification labels are 'Yes' for spoilers and 'No' for non-spoilers. Apply the following criteria rigorously: Label 'Yes' if a review: - Contains subtle references or nuanced language that hints at plot developments or character arcs, without explicit detail. - Includes emotional responses or descriptive language that indirectly reveals plot outcomes or twists. - Employs suggestive language that points to future events or endings, even if it does not reveal specific information. Label 'No' if a review: - Discusses technical aspects, acting, direction, or personal viewer impressions in a manner that does not hint at or reveal any plot details. - Comments on thematic elements, genre characteristics, or storytelling techniques without disclosing or implying crucial plot twists. - Expresses opinions on the movie's quality, performances, or execution that are free from any storyline implications or potential spoiler content. In both 'Yes' and 'No' classifications, special attention must be given to the implicit meaning behind words and the context in which they are used to prevent misclassification of subtle spoiler content and to ensure a genuine viewing experience for others. \\
 \hline
\end{tabular}

  \caption{A comparison of the prompts with the best training scores from each method on the Spoiler detection task, using 50 training samples. IPC succeeded in capturing the nuances of the GT prompt.}
  \label{tab:spoiler}
\end{center}
\end{table*}

\subsection{Generation}
\label{sct:generation}
The generation setting is composed of two parts: generating the ranker prompt and then using the ranker to optimize the generation task prompt. We tested our generation pipeline on challenging ambiguous tasks: (1) Generate a movie review that is enthusiastic, reliable and adheres to a given movie description. (2) Generate a movie review that is sarcastic but has a positive sentiment.
As in the classification case, we chose highly detailed prompts to generate the ranker GT with a scale of 1 to 5, which simulates the human preferences for the given task. For each tested method we fit a ranker using 50 labeled samples, and then optimized the generative task prompt according to the learned ranking model. The reported evaluation score is calculated by running the learned generative prompt on a test set of size 50 and evaluating the result using the target ranking prompt GT. For the baseline methods, we took samples from the IMDB review dataset~\cite{IMDB} and generated a movie description for each review. We then fed this data to the ranker optimization process. We ran both the ranker training and the generator training for 30 iteration steps.

Results for GPT-4 Turbo LLM, including both ranker training and generation prompt training, are presented in Table \ref{tab:generation}. A qualitative comparison is provided in Table \ref{tab:generation_q}. We see that using IPC improves the average ranking score of the generated reviews compared to the other tested methods in all tested scenarios. It's important to note that all the tested methods, except for IPC, performed worse than the initial prompt in some experiments. This can be explained by the distribution of the ranking scores in the real data, which is shown in Figure \ref{fig:hist_rank}, where there are almost no samples with the top score. In contrast, the distribution of the generated synthetic samples is biased towards the top two scores.

\begin{table}
    \centering
    \begin{tabular}{lccr} 
    \toprule         & Enthusiastic and reliable & Sarcastic and positive\\
         \midrule
        Initial & $4.40$\scriptsize{$\pm 0.05$} & $4.28$\scriptsize{$\pm 0.02$} \\ 
        OPRO~\cite{OPRO} & $4.31$\scriptsize{$\pm 0.02$} & $4.22$\scriptsize{$\pm 0.03$} \\
        PE~\cite{PE} & $4.09$\scriptsize{$\pm 0.09$}  & $4.76$\scriptsize{$\pm 0.09$}  \\
        IPC & \textbf{4.80} \scriptsize{$\pm 0.1$} & \textbf{4.92}\scriptsize{$\pm 0.07$}  \\
        \bottomrule
    \end{tabular}
    \vspace{+5pt}
    \caption{Average ranking score (std) of the top 5 prompts on the two generation tasks. }
    \label{tab:ablation}
\end{table}

\subsection{Ablation study}
We examine the impact of each key component of the system on the spoiler classification task. Specifically, we look at the 50 training samples case. The effect of each one of the components can be seen in Table \ref{tab:ablation}. Using synthetic data boosts model performance. It is also important to note that the analyzer component substantially improves the model's performance. This stands in contradiction to ~\cite{OPRO}'s findings that adding errors to the meta-prompts doesn't improve the model performance, and can also emphasise the gap between the standard general benchmarks and use cases such as moderation. 

\begin{table}
    \centering
    \begin{tabular}{lccr} 
    \toprule         & Spoiler classification GPT-4 Turbo & Spoiler classification GPT-3.5\\
         \midrule
        IPC (default) & $88.4$ \scriptsize{$\pm 1.7$} & $72.6$\scriptsize{$\pm 3.8$}  \\ 
        - Iterative data generation & $87.3$\scriptsize{$\pm 1.8$} & $67.0$\scriptsize{$\pm 4.1$} \\
        - Synthetic data & $83.3$\scriptsize{$\pm 2.4$}  &  $62.6$\scriptsize{$\pm 4.1$} \\
        - Analyzer & $77.8$\scriptsize{$\pm 2.2$}  & $64.3$\scriptsize{$\pm 3.8$}  \\
        \bottomrule
    \end{tabular}
    \vspace{+5pt}
    \caption{Investigation of each component's effect on the model accuracy. In each row, we removed the investigated component from the system and trained the system without it.}
    \label{tab:generation}
\end{table}

\section{Related Work}
\label{sct:related}
\textbf{Prompt Optimization.} Several methods have been suggested to address the challenge of automating the prompt engineering process. A commonly used approach is to optimize a task-specific embedding, in either a continuous~\cite{lester-etal-2021-power,li-liang-2021-prefix,DBLP:journals/corr/abs-2103-10385} or discrete~\cite{DBLP:journals/corr/abs-2302-03668,DBLP:conf/emnlp/ShinRLWS20} manner. This approach requires access to the LLM itself in order to perform the optimization. An alternative approach is to use reinforcement learning~\cite{DBLP:conf/emnlp/DengWHWGSSXH22,DBLP:journals/corr/abs-2211-11890, DBLP:conf/acl/AkyurekAKCWT23}. This approach either requires access to the generated tokens' probability or requires a large dataset for training a model. Recent works used the LLMs themselves for prompt optimization~\cite{DBLP:conf/iclr/ZhouMHPPCB23,pryzant-etal-2023-automatic,OPRO,PE}. These methods can also be applied to proprietary LLMs, , where access is limited to the final generated sentences.  However, these methods still require a good valid benchmark in order to evaluate and compare the different generated prompts, which is not always available in real-world cases.

\textbf{Synthetic data.} The utilization of synthetic data produced by LLMs has demonstrated remarkable effectiveness across a wide range of tasks, including code generation~\cite{DBLP:journals/corr/abs-2308-12950,DBLP:journals/corr/abs-2312-02418}, mathematical
reasoning~\cite{DBLP:journals/corr/abs-2308-01825,DBLP:journals/corr/abs-2308-09583}, text embedding~\cite{DBLP:journals/corr/abs-2401-00368} and text2image~\cite{BetkerImprovingIG}. The advantage of using synthetic data is not only in cost savings; it can also be beneficial for low-resource tasks or imbalanced data distributions~\cite{DBLP:journals/corr/abs-2304-13861}. Following these works, our system generates high-quality evenly distributed synthetic boundary samples, that result in a more efficient optimization process and higher-quality results.

Synthetic data was also proven to be an effective method to distil knowledge from black-box LLMs, by training on synthetic data that was generated by those models \cite{alpaca,DBLP:journals/corr/abs-2306-02707,DBLP:journals/corr/abs-2306-11644}. However, in these works the generated data was used to fully train the student model. In contrast, our work demonstrates the effectiveness of synthetic data to distil knowledge between two black-box models via automatic prompt engineering.

\textbf{Curriculum Learning.} Arranging the data samples for training machine learning models in a meaningful way, starting from easier samples and progressing to more challenging ones, can yield performance enhancements compared to the conventional method of training based on random data shuffling. This approach is known as curriculum learning~\cite{curic,curic2}. Curriculum Learning has been proven to be effective in various fields such as object localization~\cite{DBLP:conf/cvpr/IonescuALPPF16,DBLP:conf/eccv/ShiF16}, object detection~\cite{DBLP:conf/iccv/ChenG15,DBLP:journals/pami/SanginetoNCS19} and NLP~\cite{DBLP:conf/ranlp/KocmiB17,DBLP:conf/naacl/PlataniosSNPM19}. Inspired by these ideas, in~\cite{DBLP:journals/corr/abs-2401-01335} they propose to fine-tune LLMs by iteratively generating synthetic data and refining the policy to distinguish between the synthetic data and the human-annotated data. In our work, we use a similar approach, where the system iteratively generates more challenging cases that resolve the previous prompt ambiguity in order to more efficiently tune to the user intent.
\section{Conclusions}
In this work, we introduced IPC, a system for automatic prompt engineering. The system combines a synthetic data generation module that generates challenging and diverse samples, and a prompt optimization module that suggests new prompts. Both of them are implemented by prompting LLMs, and they iteratively refine each other until the prompt converges. We further propose a new method to extend the meta-prompt based prompt optimization process to generative tasks. We demonstrate the effectiveness of our system on real-world use cases such as moderation and generation with respect to strong proprietary models (GPT-3.5/4-Turbo).Our method significantly enhances the resulting performance of prompts in all tested scenarios.

Our system is built in a modular and flexible way that allows for easy modification and addition of new components. In future work, we intend to extend our system to new use cases such as multi-modality and in-context learning. We also intend to explore further possibilities to optimize the meta-prompts themselves.

\bibliography{neurips_2023.bib}

\newpage
\appendix
\addcontentsline{toc}{section}{Appendix}
\part{Appendix} 
\section{Implementation details}

\definecolor{codegreen}{rgb}{0,0.6,0}
\definecolor{codegray}{rgb}{0.5,0.5,0.5}
\definecolor{codepurple}{rgb}{0.58,0,0.82}
\definecolor{backcolour}{rgb}{0.95,0.95,0.92}

\lstdefinestyle{mystyle}{
    backgroundcolor=\color{backcolour},   
    commentstyle=\color{codegreen},
    numberstyle=\tiny\color{codegray},
    basicstyle=\ttfamily\footnotesize,
    breakatwhitespace=false,         
    breaklines=true,                 
    captionpos=b,                    
    keepspaces=true,                 
    numbers=left,                    
    numbersep=5pt,                  
    showspaces=false,                
    showstringspaces=false,
    showtabs=false,                  
    tabsize=2
}

\lstset{style=mystyle}

\begin{figure}
  \centering
  \includegraphics[width=1.\linewidth]{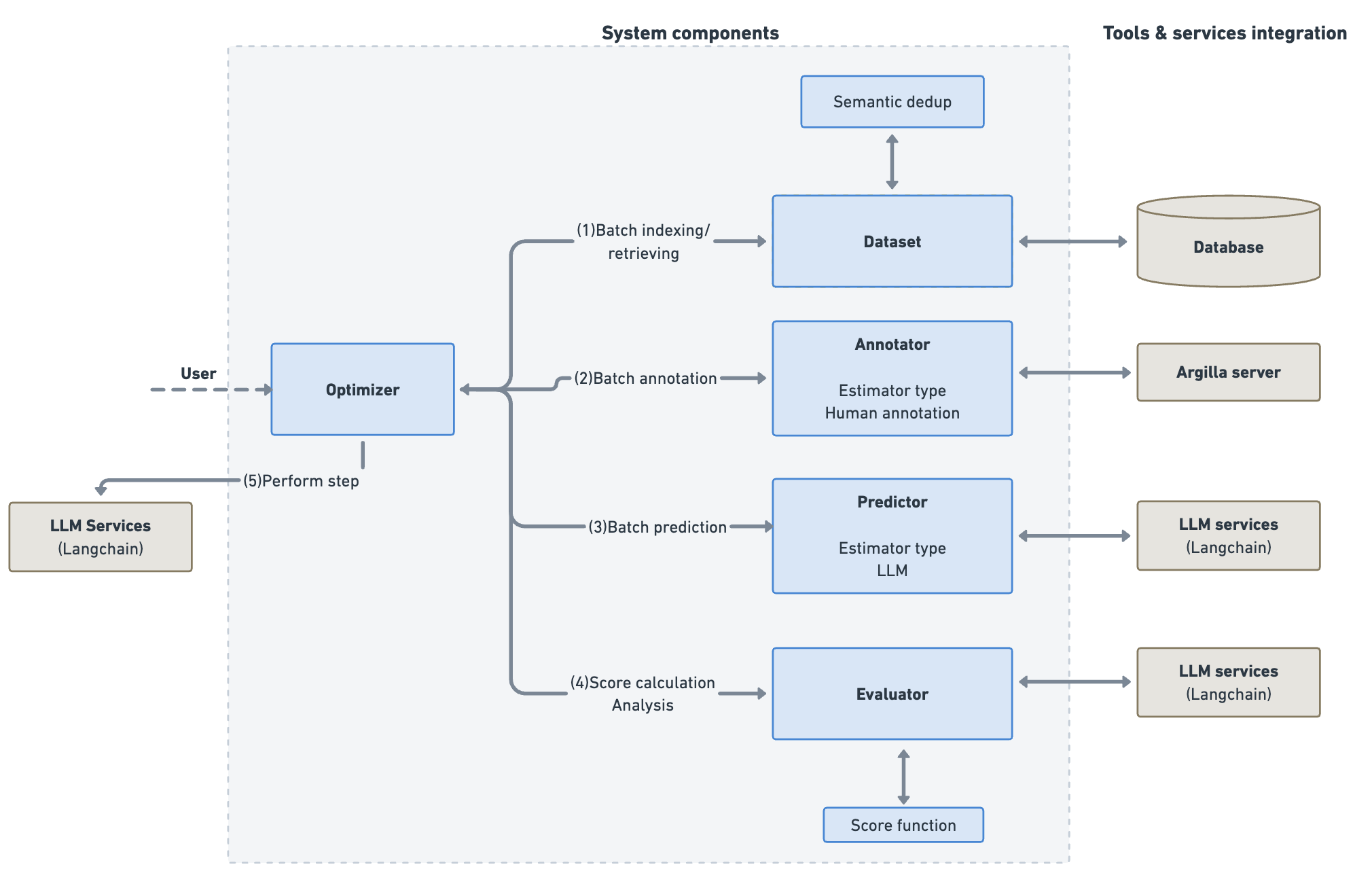}
  \caption{Architecture overview.}
  \label{fig:arch}
\end{figure}

In this section, we provide a additional information on the implementation details.
An architecture overview of our system is provided in Figure~\ref{fig:arch}.
We also provide a list of the meta-prompts utilized within the pipeline of our system. 

\vspace{0.5cm}

\texttt{Generating initial samples (iteration 0)}
\begin{lstlisting}[language=Python]
Assistant is a large language model designed to generate challenging samples for every task.
Generate a list of {num_samples} challenging samples for the following task.
### Task description:
{task_description}
### Task Instruction:
{instruction}
###
The generated samples should be challenging and diverse such that using the task instruction as a prompt will result in the wrong result.
The number of generated samples from each class should be balanced (i.e. the same number of samples for each class)
\end{lstlisting}

\texttt{Generating samples (iteration >0)}
\begin{lstlisting}[language=Python]
Assistant is a large language model designed to generate challenging samples for every task.
Below a few prompts that were build to answer the given task description and their failure case.
Task description:
{task_description}

## Examples of common failure, each sample is followed by the the model prediction and the GT (ground truth)
{history}
######
Here are few unique samples derived from realistic scenarios for the task outlined above.
## Realistic Samples
{extra_samples}
#####
This was the new proposed prompt:
## Prompt
{prompt}

Your task is to generate {num_samples} by following this guidelines:
1. The generated samples should be diverse
2. They should preserve the style and the length of the given examples
3. The samples must be challenging and hard to classify by the model. This can be achieved by:
    1. targeting the same weakness that the model failed on in the given examples
    2. targeting weakness that are different from the existing examples in the failure cases
4. The number of generated samples from each class should be balanced (i.e. the same number of samples for each class)
\end{lstlisting}

\texttt{Analyzer prompt}
\begin{lstlisting}[language=Python]
Assistant is a large language model designed to provide a high quality analysis for every task.
You are given the following task description
{task_description}

Here is the prompt instructions that was given to the model:
{prompt}

An expert ranker evaluated the model's performance on the given task description.
and rank according to the following scale: {labels}

The mean score for this prompt is: {accuracy}
##
Here is a list of challenging cases for the given prompt and their rank:
##Challenging Cases:
{failure_cases}

###
Note that the ranker labels are __absolutely correct__, but the prompts (task descriptions) may be incorrect and need modification.
Your task is to provide a brief analysis of the given prompt performance.
Guidelines:
1. The analysis should contain only the following information:
    - A summary of the common mistakes of the prompt and the ways he can be improve his generation, try to cluster the failure cases into groups and describe each group.
2. The total length of your analysis should be less than 200 token!
###
Analysis:
\end{lstlisting}

\texttt{Generating new proposed prompt}
\begin{lstlisting}[language=Python]
Assistant is a large language model designed to provide the best prompt for every task.
Below are a few suggested prompts for the task and their score, for the following task:
{task_description}

## Examples
{history}
######
This is the error analysis for the last prompt:
{error_analysis}
######
Your task is to generate:
1. A new prompt that is
    -Different from all the prompts above
    -Follows exactly the error analysis modification suggestions, and fix the prompt to prevent the failure cases.
    -Has a higher score than all the prompts above.
2. The predicted score of this prompt

You must adhere the error analysis instructions! even in case it seems there is a contradiction between these instructions, and the task. The error analysis is tested on a ground truth, thus represent the exact intent of the task.
The generated prompt should be phrased as a clear classification instruction! it should not include any instructions and descriptions on the modification that should be done to the prompt.
Note that the previous prompt contains an implicit assumptions on the intent of the task that might be incorrect. You should replace this assumption with more accurate assumptions using the score of the previous prompts and the error analysis.
The result prompt should indicate that the task is a classification class with the following labels {labels}!
\end{lstlisting}

\texttt{Modifying the task description for training the ranker}
\begin{lstlisting}[language=Python]
Assistant is a large language model designed to generate a task description.
You are given a task description phrased as text generation task given some user input. Your task is to rephrase it as a task that suppose to evaluate the quality of the given generative task and how well it adhere to the user input.
#####
Input task description: {task_description}
#####
Rephrased task description:
\end{lstlisting}

\texttt{Modifying the initial prompt for training the ranker}
\begin{lstlisting}[language=Python]
Assistant is a large language model designed to generate instructions for every task.
You are given a instructions phrased as text generation task.
Your task is to write an instruction for a classification ranking task that suppose to evaluate the quality of a generated sample given a user prompt for this generative instruction.
Guidelines:
1. The classifier labels are {label_schema}. The result instructions should indicate explicitly that the task is a classification class with the following labels {label_schema}!
2. The generated instruction must also evaluate how well the generated sample adhere the user prompt
#####
Input generative instruction: {prompt}
#####
Rephrased classification quality evaluation instruction:
\end{lstlisting}

\section{Experiments: Additional details}
In this section, we provide additional material on the experiments provided in the paper.

\begin{table*}[!h]
\begin{center}
\begin{tabular}{p{2.5cm}| p{11cm}} 
\shline

 GT ranker  & Establish a revised five-point assessment scale for the movie review generator that emphasizes precise differentiation between all levels of reflective accuracy and expressed enthusiasm. Assign the classification labels "1", "2", "3", "4", and "5", with each level distinctly representing the depth and sincerity of reflection on the movie description as well as the intensity and authenticity of enthusiasm:
1. Deficient Reflection and Lacking Enthusiasm ("1"): Reviews that fundamentally misrepresent or ignore the movie description and display a lack of genuine enthusiasm, signaling a complete disconnect and an ineffective critique.
2. Basic Reflection and Low Enthusiasm ("2"): Reviews that merely mention aspects of the movie description without depth and show only a low level of enthusiasm, resulting in a critique that lacks persuasive power and engagement.
3. Adequate Reflection and Moderate Enthusiasm ("3"): Reviews that reflect the movie description accurately and convey moderate enthusiasm, offering a fair and constructive critique that is neither overly zealous nor dispassionate.
4. Detailed Reflection and High Enthusiasm ("4"): Reviews that capture the nuances of the movie description in detail and exhibit high enthusiasm, presenting a critique that is both engaging and substantively insightful.
5. Exceptional Reflection and Intense Enthusiasm ("5"): Reviews that deeply engage with the movie description, demonstrating both an exceptional level of reflection and intense enthusiasm, with critiques that significantly enhance the analysis and captivate the audience. 
This adjusted scale is designed to correct the prior misclassifications by ensuring clear and accurate recognition of each level of performance, thereby avoiding the underestimation of high-quality reviews and providing a fair representation for lower-quality outputs.
\\ 
 \hline
 \\ 
 Initial Prompt

 \textbf{(Mean rank: 4.4)}
  & Generate an authentic enthusiastic movie review for the following movie \\ 
 \hline
\\ 
OPRO~\cite{OPRO}

 \textbf{(Mean rank: 4.3)}& Compose a compelling and positive movie review for a recent release, capturing the film's essence while highlighting unique elements that make it stand out. \\
 \hline
 \\ 
 PE~\cite{PE} 
 
  \textbf{(Mean rank: 4.1)}& Create a movie review generator that articulates sophisticated genre-specific language while ensuring that advanced terminology is immediately followed by comprehensible explanations. This approach guarantees that the review remains accessible to a broad readership, including both seasoned cinephiles and those new to film criticism. The generator must offer an in-depth narrative analysis, employing narrative theory to enrich the commentary on the film's plot, character arcs, and thematic substance, all while maintaining an educational yet non-academic tone. It should craft reviews that exude excitement and provide an even-handed critique, spotlighting the film's artistic accomplishments as well as constructively addressing areas for improvement. The result should be an authoritative and engaging review that captures the essence of the movie with enthusiasm and critical precision, appealing to a wide spectrum of movie enthusiasts.  
 \\
 \hline
 \\
 IPC
 
 \textbf{(Mean rank: 4.8)}& Devise a movie review generator that excels in providing a balanced and nuanced critique, ensuring that each review exudes genuine enthusiasm for the movie's triumphs while offering an insightful, yet subdued, evaluation of its less successful elements. The generator must consistently utilize precise and vibrant examples to underpin its praises, such as remarkable performances, poignant plot developments, and groundbreaking directorial choices. It is vital to include a section in the review that thoughtfully places the film within its genre and historical cinematic context, highlighting its unique contributions and potential influence. Reviews should read as an authoritative, persuasive endorsement of the film's merits, inviting readers to not only enjoy but also understand its significance in the film world, all while keeping the tone wholly positive and engaging.  \\
 \hline
\end{tabular}

  \caption{A comparison between the prompt with the best training score in each method on the authentic and enthusiastic generative task. The GT ranker is provided in the first row. While OPRO and PE result in a prompt that performs worse than the initial simple prompt. Using IPC results in a significantly better prompt.}
  \label{tab:generation_q}
\end{center}
\end{table*}

\begin{table*}[th]
\begin{center}
\begin{tabular}{p{3.5cm}| p{9cm}} 
\shline
\\ 
 Sentiment GT & Does this movie review have a positive sentiment? Answer Yes or No. Answer No only in case there is an explicit expression of dissatisfaction otherwise, answer yes.
\\ [0.5ex] 
 \hline
 \\ 
 Sentiment Initial Prompt 
  & Does this movie review have a positive sentiment? answer Yes or No \\ 
 \hline
\\ 
PG GT& Augment the IMDB review classifier to better recognize and classify reviews containing adult themes. The classifier should label a review as 'Yes' when it detects not only explicit descriptions of sex, violence, and strong language but also subtle adult themes, such as suggestive dialogue, nuanced portrayals of relationships, and indirect references to sex or desire. Additionally, the classifier should account for adult humour and jokes that are presented in a lighthearted manner, ensuring they are not missed. It should also accurately identify implicit adult content conveyed through cinematic techniques or complex narratives, as well as non-graphic mature content like restrained portrayals of infidelity or taboo relationships. Furthermore, personal remarks on the attractiveness of actors or suggestive comments that reviewers make must be included in the 'Yes' category. Conversely, reviews that lack explicit adult content, significant implications of mature themes, and any aforementioned subtleties should be classified as 'No'.
 \\
 \hline
 \\ 
 PG Initial Prompt 
  & Does this movie contain ah adult content? answer Yes or No
 \\
 \hline
\end{tabular}

  \caption{Prompts for the PG and sentiment classification tasks.}
  \label{tab:all_experimants}
\end{center}
\end{table*}

\begin{figure}[h]
    \centering
    
    \begin{minipage}[b]{0.49\linewidth}
        \centering
        \includegraphics[width=\linewidth]{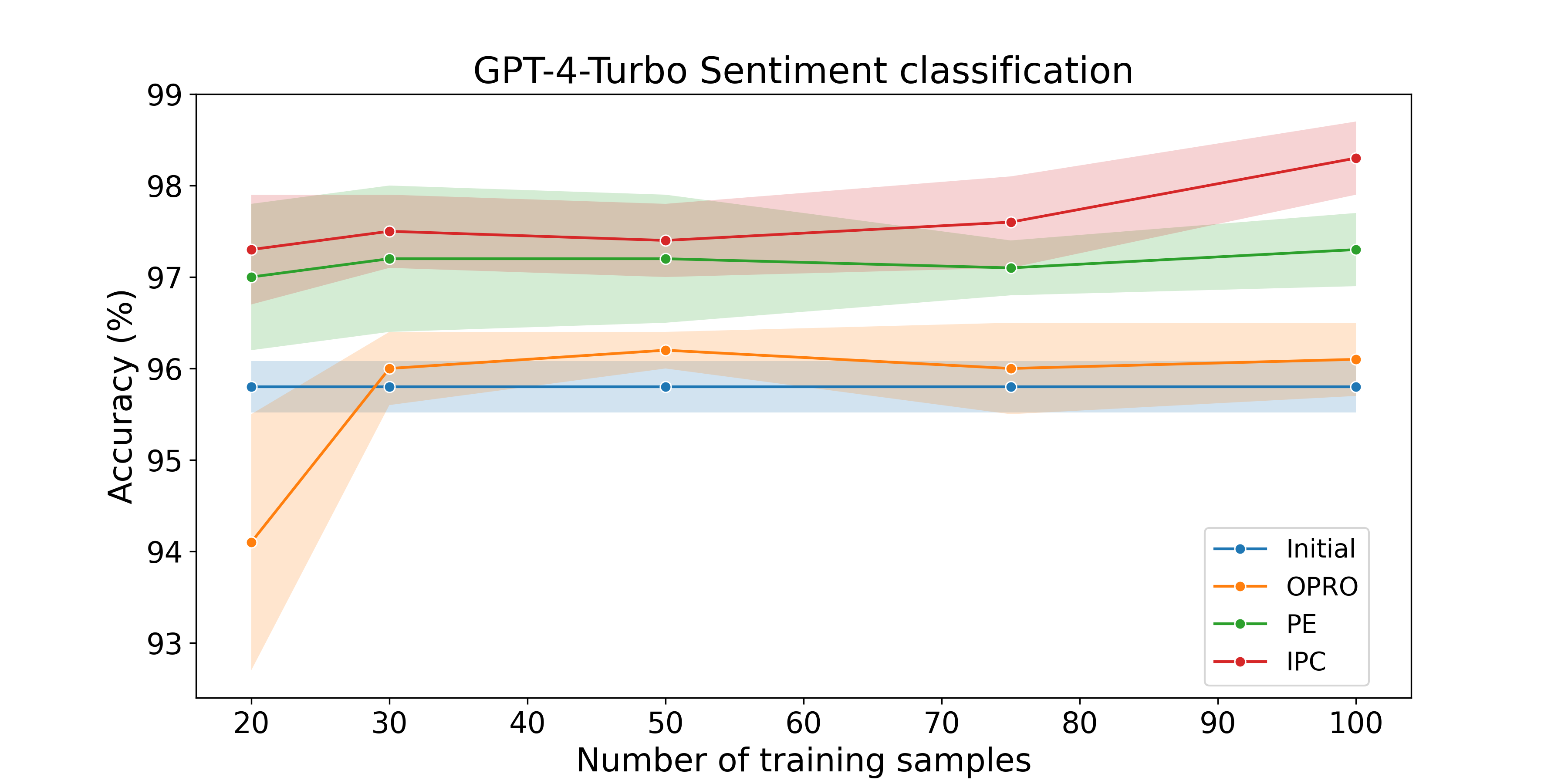}
        \label{fig:image3}
    \end{minipage}
    \hfill
    \begin{minipage}[b]{0.49\linewidth}
        \centering
        \includegraphics[width=\linewidth]{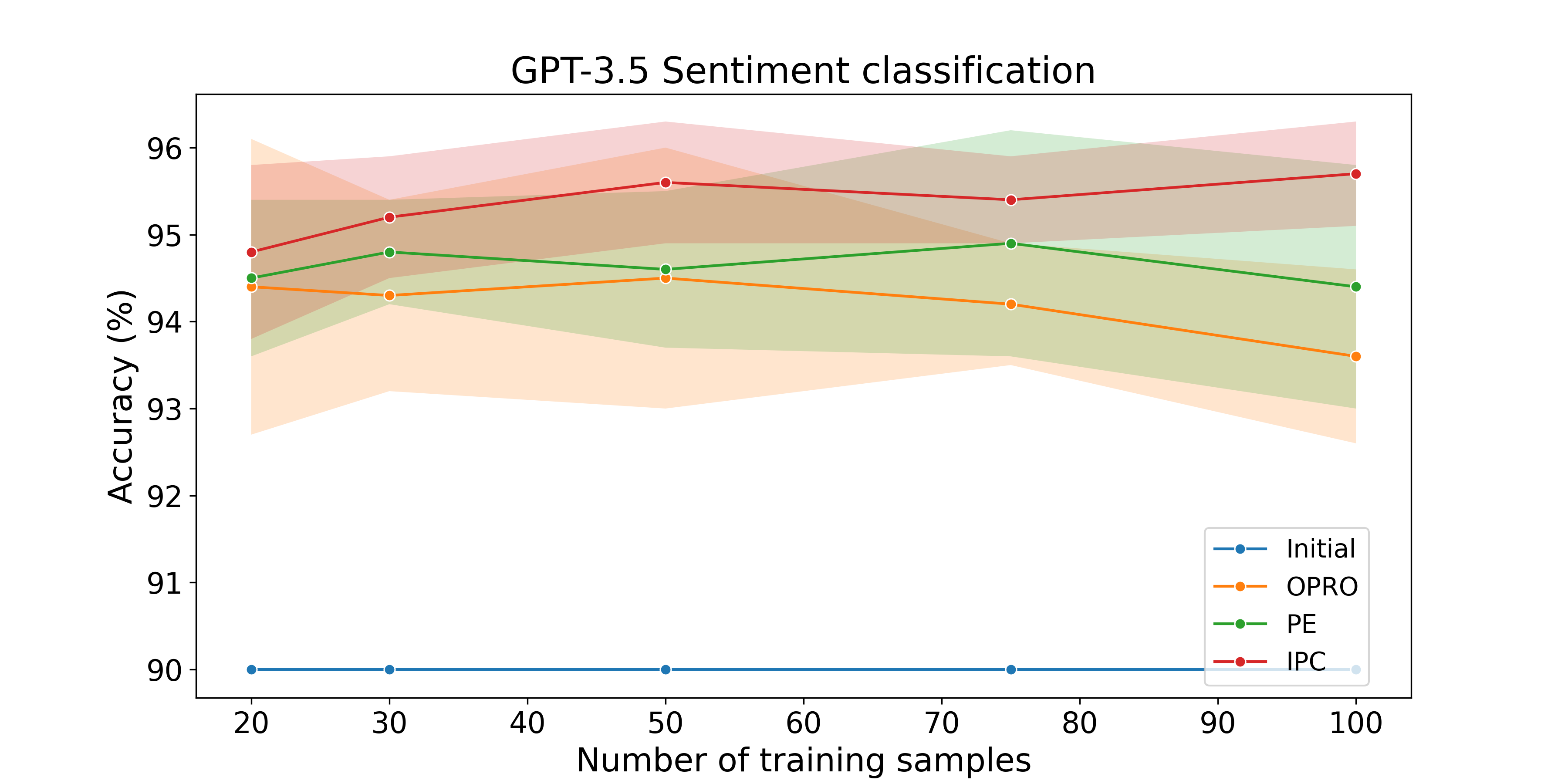}
        \label{fig:image4}
    \end{minipage}
    \label{fig:all_images}

    \caption{Accuracy on the sentiment classification task, with respect to different numbers of training steps. IPC outperforms other tested methods, and results in lower variance.}
    \label{fig:all_images}
\end{figure}

\end{document}